\documentclass[conference]{IEEEtran}
\IEEEoverridecommandlockouts
\usepackage{cite}
\usepackage{amsmath,amssymb,amsfonts}
\usepackage{algorithmic}
\usepackage{graphicx}
\usepackage{textcomp}
\usepackage{xcolor}
\def\BibTeX{{\rm B\kern-.05em{\sc i\kern-.025em b}\kern-.08em
    T\kern-.1667em\lower.7ex\hbox{E}\kern-.125emX}}

\usepackage{caption}
\usepackage{subcaption} 
\usepackage{graphicx}
\usepackage{multirow}
\usepackage{url}
\usepackage{hyperref}
\usepackage{flushend}
    
\begin{document}

\title{GNN-Ensemble: Towards Random Decision Graph~Neural Networks
}


\author{\IEEEauthorblockN{Wenqi Wei\IEEEauthorrefmark{1}\IEEEauthorrefmark{3}\IEEEcompsocitemizethanks{\IEEEauthorrefmark{3}This Work was done when the author was with IBM Research.}, 
Mu Qiao\IEEEauthorrefmark{2}, 
Divyesh Jadav\IEEEauthorrefmark{2} 
}

\IEEEauthorblockA{\IEEEauthorrefmark{1}Department of Computer and Information Sciences, Fordham University, New York, NY, USA}

\IEEEauthorblockA{\IEEEauthorrefmark{2}IBM Research - Almaden, San Jose, CA, USA}

\IEEEauthorblockA{Email: wenqiwei@fordham.edu, \{mqiao,divyesh\}@us.ibm.com}
}

\maketitle

\begin{abstract}
Graph Neural Networks (GNNs) have enjoyed wide spread applications in graph-structured data. However, existing graph based applications commonly lack annotated data. GNNs are required to learn latent patterns from a limited amount of training data to perform inferences on a vast amount of test data. The increased complexity of GNNs, as well as a single point of model parameter initialization, usually lead to overfitting and sub-optimal performance. In addition, it is known that GNNs are vulnerable to adversarial attacks. In this paper, we push one step forward on the ensemble learning of GNNs with improved accuracy, generalization, and adversarial robustness. Following the principles of stochastic modeling, we propose a new method called GNN-Ensemble to construct an ensemble of random decision graph neural networks whose capacity can be arbitrarily expanded for improvement in performance. The essence of the method is to build multiple GNNs in randomly selected substructures in the topological space and subfeatures in the feature space, and then combine them for final decision making. These GNNs in different substructure and subfeature spaces generalize their classification in complementary ways. Consequently, their combined classification performance can be improved and overfitting on the training data can be effectively reduced. In the meantime, we show that GNN-Ensemble can significantly improve the adversarial robustness against attacks on GNNs.

\end{abstract}



\section{Introduction}

Graph is widely used to model many real-world relationships, 
ranging from social and rating networks~\cite{newman2002random}, financial transactions~\cite{wei2020bitcoin}, healthcare~\cite{fu2022hint}, to protein and molecule networks~\cite{hamilton2017inductive,fu2021differentiable}.
Graph neural network (GNN)~\cite{bruna2013spectral,kipf2016semi} has become the mainstream
methodology for deep learning on graphs. GNN
has shown
promising performance on various applications on graph data, for example, identifying components in protein molecules by node classification, anomaly detection for financial transactions with link prediction, and interest group finding with community detection. Through GNNs, 
each node in the graph builds its representation based on its node feature as well as the features of its neighbors through message passing. The representation of the node is updated using an aggregation of these messages. 

While GNNs can obtain reasonably good accuracy on many graph learning tasks, their increased representational capacity comes from higher model complexity. This can lead to overfitting and therefore weakens the generalization ability. In both transductive and inductive settings, graph data are typically with a small amount of labeled training data and a large amount of unlabeled test data. A trained GNN may easily overfit the small training data. In the meantime, although the current GNN models utilize the neighborhood sampling of the target node and aggregate such neighborhood information as its representation, a single trial of model parameter initialization for deep learning could converge to local optima. DropOut~\cite{srivastava2014dropout} is a popular regularization technique for deep learning models to avoid overfitting. However, in GNNs, \cite{kipf2016semi}
indicates that DropOut alone is ineffective in preventing overfitting. This is partially due to the fact that when GNN becomes very deep, the Laplacian smoothing will make all nodes’ representations converge to a stationary point such that they are no longer related to node features~\cite{li2018deeper}. 

Additionally, like other deep learning methods, GNNs are vulnerable to adversarial attacks~\cite{zugner2018adversarial}. 
If the attacker maliciously adds or drops edges concerning a target victim node, the prediction related with that node may change. For example, a malicious merchant may link her products to specific items in order to fool the graph recommendation systems for higher exposure. In another example, a malicious user may manipulate her profile to build connections with 
targeted users to mislead the friend recommendation system. Such kind of vulnerability poses serious threats to GNNs, especially under safety-critical scenarios.

Ensemble learning has been widely used to boost the generalization and training stability
with the combined wisdom ~\cite{breiman1996bagging,breiman2001random}.
Deep neural network ensembles have gained increasing popularity in the deep learning community for improving the accuracy performance and robustness~\cite{wu2021boosting,pang2019improving,wei2020robust}. It is equally important to improve the accuracy performance of the GNN models as well as their robustness against adversarial attacks. In this paper, we aim to shed some insight into the following research questions: (1)
Can we utilize an ensemble of GNN models, despite their different local optimal results, to achieve better classification performance? (2) Can we generate GNN ensembles with better generalization power and reduce overfitting? (3) Can we increase the robustness of GNN against adversarial attacks via GNN ensemble? 

In this paper, we present GNN-Ensemble, an ensemble learning approach that combines multiple GNN models trained with both randomly selected substructures in the topological space and randomly selected subfeatures in the feature space. The development of GNN-Ensemble consists of three main steps. 
First, we subsample the graph structure for subgraphs and the node features for subfeatures, assuming the graph is attributed. 
Second, we train multiple base GNN models, e.g., GraphSage~\cite{hamilton2017inductive} with different sets of subgraphs and subfeatures for independent decision making. 
The third and final step is to aggregate multiple GNN decisions with ranking or voting. By the theory of stochastic modeling with a discriminant function for joint decision making~\cite{ho1995random}, these GNNs in different substructure and subfeature spaces generalize their classification in complementary ways. 
Consequently, their combined classification can be monotonically improved and overfitting on the training data can be reduced. With experimental results on four benchmark graph datasets\footnote{\url{https://pytorch-geometric.readthedocs.io/en/latest/modules/datasets.html}} (two for small graphs Cora and PubMed, and two for large graphs PPI and Reddit), we demonstrate that the proposed GNN-Ensemble method is general and can be used with many other backbone GNN models. GNN-Ensemble improves the accuracy performance, reduces overfitting, and improves the robustness against adversarial attack when compared to a single GNN model. 

The rest of the paper is organized as follows. Section II reviews the related work in GNNs, adversarial attacks and defense on the graph, and ensemble learning.
In Section III, we provide a preliminary of the problem formulation and graph neural networks. In Section IV, we present the systematic creation of the GNN-Ensemble method with graph structure and node feature sampling and a discussion on diverse model generation. We further define the discriminant function, followed by the rules on decision aggregation and complexity analysis. Experimental results are reported in Section V. We finally conclude in Section VI.

\section{Related Work}

\textbf{Graph Neural Networks.} GNNs apply deep learning to graph data by capturing local graph structure and feature information in a trainable fashion to derive powerful node representations~\cite{scarselli2008graph}. Over the past few years, GNNs have achieved great success in solving machine learning problems on graph data. Under the umbrella of Graph Convolutional Networks (GCNs), there are two main families of GNNs for graph convolution, i.e., spectral methods and spatial methods. For the former, \cite{bruna2014spectral} uses the Fourier bases of a given graph for graph convolution. \cite{defferrard2016convolutional} utilizes Chebyshev polynomials as the convolution filter. ~\cite{kipf2016semi} proposes GCN based on the first-order approximation of spectral graph convolution. For the latter, graph convolution aggregates and transforms local information of a given node in the spatial domain. GraphSage~\cite{hamilton2017inductive} learns aggregators by sampling and aggregating neighbor information. FastGCN~\cite{chen2018fastgcn} proposes an ad hoc sampling scheme to restrict the neighborhood size. ClusterGCN~\cite{chiang2019cluster} restricts the neighborhood search within a subgraph identified by a graph clustering algorithm. \cite{chen2018stochastic} restricts neighborhood size by requiring only two support nodes in the previous layer. Graph Attention NeTwork (GAT)~\cite{velivckovic2017graph} introduces attention mechanism for neighbor information aggregation. GraphSaint~\cite{zeng2019graphsaint} preprocesses the training graph by subsampling, similar to our substructure sampling for base model training. However, the subgraph sampling in GraphSaint serves as the bootstrap sampling for training a single GCN model while our subgraph sampling is used for bagging. For a thorough review, we refer readers to recent surveys~\cite{wu2020comprehensive,zhou2022network}.

Overfitting reduction of GNNs is usually discussed with over-smoothing reduction~\cite{li2018deeper}. DropEdge~\cite{rong2019dropedge} removes edges uniformly as a data augmentation technique for node classification. GraphCrop~\cite{wang2020graphcrop} crops the contiguous subgraphs from the original graph to generate novel and valid augmented graphs. \cite{hasanzadeh2020bayesian} introduces DropOut as a Bayesian approximation and uses Monte Carlo estimation of GNN outputs to evaluate the predictive posterior uncertainty.
While these approaches focus on enabling deep GCN with more layers by reducing over-smoothing and gradient vanishing, our work aims to reduce overfitting on training data to improve the node classification performance of existing GNN methods. Our method is motivated by the relation between decision tree and random forests.

\begin{figure*}[t]
  \centerline{\includegraphics[scale=.52]{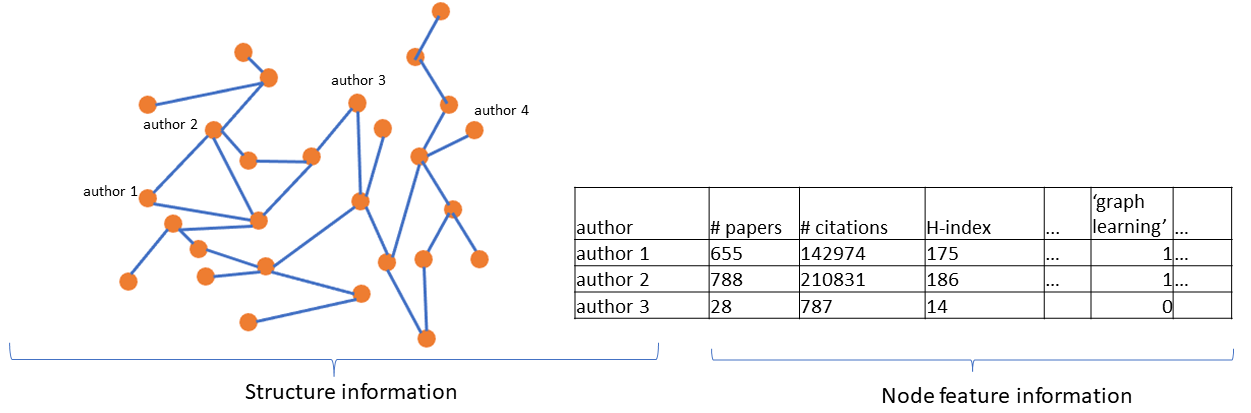}}
\caption{Topological and node attribute information.}
\label{fig:graph_info} 
\end{figure*}

\textbf{Ensemble Learning.} Ensemble learning has been widely adopted to improve the accuracy, stability, and robustness of machine learning systems.
The main idea of ensemble learning is to construct a set of individual machine learning models (i.e., base model) and combine their predictions to yield better performance. 
Ensemble learning can be categorized into three threads. The first trains tens or hundreds of weak models for decision making, such as bagging~\cite{breiman1996bagging} and  boosting~\cite{freund1996experiments}. 
For example, random forests~\cite{breiman2001random} is the most influential bagging approach while XGBoost~\cite{chen2015xgboost} is the most well-known boosting method. 
The dropout operation and layerwise feedforward process of neural networks benefit from such decision ensemble with improved accuracy and reduced overfitting. The second aims to identify high-quality ensembles from a pool of strong deep learning base models with high individual model accuracy and different neural network backbones~\cite{wu2021boosting}. Since different deep learning models converge to different local minima~\cite{hansen1990neural}, ensembles of diverse high-quality deep neural network are typically used to improve accuracy and adversarial robustness~\cite{pang2019improving,wei2020robust}. 
The third centers on voting methods during decision aggregation, such as simple averaging, majority voting, and plurality voting.
Existing GNN ensembles leverage multiple GNN learners for domain-specific applications. \cite{kosasih2021graph} trains multiple GNNs on molecular graphs for molecular property prediction. \cite{chakravarty2020learning} models the comorbidity statistics from ensembles of CNN models as a graph and uses GNN for chest radiograph screening. The most related work to ours leverages the idea of adaboost~\cite{sun2020adagcn} for deeper graph learning. Different from all the existing works, our proposed GNN-Ensemble integrates all three threads of efforts in ensemble learning. We sample substructure and subfeature of the original graph and train hundreds of base models for decision making using the bagging method. Different voting methods can be exploited in GNN-Ensemble during the decision aggregation phase. Our method is further open to optimizing with diverse base model selection for high-quality ensemble. 

\textbf{Adversarial attacks and defenses on the graph.} GNNs are known to be vulnerable to adversarial attacks. The adversary can mislead GNNs to make wrong predictions by manipulating a few edges and nodes or modifying the node features. Adversarial attacks on the graph can be performed at the training phase as poisoning~\cite{zugner2018adversarial_meta,bojchevski2019adversarial} or at the prediction phase as evasion~\cite{dai2018adversarial,xu2019topology,chang2020restricted,wu2019adversarial,zugner2018adversarial}. We refer readers to the latest survey for a comprehensive overview~\cite{jin2003adversarial}. Since adversarial robustness is considered as
an efficient tool to analyze both the theoretical properties as well as the practical accountability~\cite{liu2018benchmarking}, we draw our attention to the robustness against adversarial attacks on the graph in addition to the accuracy performance. We focus on adversarial attack mitigation at the inference phase.

\section{Preliminary}

Our method draws the theoretical foundation from random forests~\cite{ho1995random}. The discriminant power from stochastic modeling in random forests seeks to solve a classification problem in a given feature space. Therefore, we need to identify the information needed for constructing the base model. Then we provide an overview of how a single base GNN model utilizes these information for graph learning.

\subsection{Graph Information }

\textbf{Two types of graph information.} Given a graph $G = (V, E)$ with the node set $V$ and the edge set $E$, an attributed graph contains two types of information, i.e., structure information which describes the connectivity of nodes, and node feature information which describes the attributes of 
nodes. The edge connections can be represented by an adjacency matrix: $A \in \{0,1\}^{n \times n}$, where $n=|V|$, the total number of nodes. Let $d$ denote the dimension of node features.  
The node feature matrix is denoted as $F \in \mathbb{R}^{n \times d}$. 
\textbf{Figure~\ref{fig:graph_info}} illustrates the two types of graph information in an academic collaboration graph. On the left side, each author is represented as a node, and the collaboration relationship is modeled as an edge. The table on the right side lists the attributes of each author, such as
the number of published papers, number of citations, and research interests.

\textbf{Node classification.} We focus on the task of node classification in this paper. Given a graph with some labeled nodes, the classification task predicts the class for those unlabeled nodes. There are two kinds of settings in node classifications, i.e., transductive and inductive settings.

\begin{itemize}
    \item The transductive setting observes all the data beforehand, including both the training and test datasets. We learn from the already labeled training data and then predict the node labels of the test data. Even though we do not know the labels of the test data, we can still leverage the graph connections and additional information present in the test data during the training process.
    \item The inductive setting is about learning general rules from observed training data. The test data is disjoint from the training data and unseen during the training process. 
    The classification model is trained with only the training dataset. The trained model is then applied to predict the labels of a test dataset.
\end{itemize}

\subsection{Graph Neural Networks} 
Given a graph $G = (V, E)$, a GNN takes the following two types of information as input: (1) an input feature matrix $F$, and (2) an adjacency matrix $A$.
The forward propagation rule in GNNs determines how the information from the input will go to the output side of the neural network. We will cover the commonly used shallow-layer GNNs, which typically have three layers: the input layer, the hidden layer for latent representation, and the output layer. Given a node $v$, its representation at each layer can be modeled as:
$$h_v^l=\sigma\bigg(\mathcal{W}_l\sum \nolimits_{u\in\eta(v)}\frac{h^{l-1}_u}{\eta(v)}+\mathcal{B}_l h_v^{l-1}\bigg).$$
where $l=1,..L$ is the layer number, $\sigma$ is the ReLU activation function, and $\eta(v)$ is the neighborhoods of node $v$. $\mathcal{W}$ and $\mathcal{B}$ are the weight and bias of the GNN model, respectively. The first part on the right side averages all the neighbors of node $v$ and the second part is the previous layer embedding of node $v$ multiplied by a bias $\mathcal{B}_l$. Then the output will be the embedding after $L$ layers of neighborhood aggregation. GraphSage~\cite{hamilton2017inductive} modifies the function of this simple neighborhood aggregation as
$$h_v^l=\sigma\bigg(\big[\mathcal{W}_l \cdot AGG\big(\{h_u^{l-1}, \, \forall u \in \eta(v)\}\big), \mathcal{B}_l h_v^{l-1}\big]\bigg),$$ 
where AGG denotes the aggregation function. 
Other GNN models used in this paper, e.g., FastGCN~\cite{chen2018fastgcn}, ClusterGCN~\cite{chiang2019cluster}, GraphSaint~\cite{zeng2019graphsaint}, GAT~\cite{velivckovic2017graph}, introduce graph convolution layer, graph sampling, attention mechanism, and alternative techniques to refine the simple neighborhood aggregation rule for more generalized and advanced graph learning.

\section{Systematic Creation of GNN Ensemble}

In this section, we will present the proposed GNN-Ensemble method. We will first introduce the substructure sampling and subfeature sampling, followed by a discussion on the model diversity. Then, we will define the discriminant function and discuss the decision aggregation rules. Finally, we will analyze the time and space complexity for GNN-Ensemble. 

\subsection{GNN Ensemble}


\textbf{Substructure and Subfeature Sampling.}
The underlying idea behind GNN-Ensemble is to first generate many base models and then combine these base models  to form a new stochastic model. The key point is that the ``projectability'' of the combined stochastic model will be comparable to that of the base models from which it is built, where the projectability is defined as 1 minus the difference in performance between training and test (the larger the projectability, the more projectable the model)~\cite{kleinberg1996overtraining}. Our method to create multiple GNN base models is to construct GNNs in randomly selected subspaces of the feature space as well as randomly selected subgraphs of the topological space.
For a given topological space with $n$ number of nodes, $m$ number of edges, and feature space of $d$ dimensions, there are in total $2^{(n+m+d)}$ subspaces in which a base GNN can be constructed. We use randomization to introduce differences into classifiers, i.e., randomly select the substructures and subfeatures.

The base GNN model is trained in each selected substructure and subfeature space. Each of these base GNN models classifies the training data with high accuracy, aiming to discern between the inputs of different classes while not discerning between training and test data of the same class~\cite{kleinberg1996overtraining}. Consequently, the classification is invariant for points that are different from the training points only in the unselected dimensions~\cite{ho1995random}. Thus each base GNN generalizes its classification in a different way. The vast number of substructures and subspaces provides more choices than can be used in practice. The performance on the training set of the stochastic model can be made arbitrarily good if enough base models are used. 

With substructure and subfeature sampling, we generate different random subgraphs out of the original graph. Such randomness and diversity of the input graphs can serve as a data augmentation approach. In addition to the understanding of ensemble from random forests, we provide another intuitive understanding of GNN-Ensemble from the perspective of graph learning. The key in the GNN model is to aggregate neighbors’ information for each node as a weighted sum of the neighbor features (the weights are associated with the edges). Those approaches used for neighborhood sampling in GNNs, such as GraphSage and ClusterGCN, enable a random subset aggregation instead of the full aggregation during GNN training. In GNN-Ensemble, when GraphSage and ClusterGCN are used as the backbone models, the neighbor aggregation is more sparse. Therefore, the trained GNN model is less likely to remember random error patterns encoded in the training data.

\textbf{Model Diversity.} The randomness in the selected substructure and subfeature enables the initialization of training algorithms with different input data, and therefore eventually yields different classifiers. Similar to the performance of decision trees in random forests, the base GNN model can also provide very high accuracy on training data. This may lead to overfitting~\cite{geman1992neural}, although much reduced compared to the single GNN model and GNN ensemble with each model trained with full sets of the structure and feature space. The best scenario is when all base GNNs of an ensemble can learn and predict with uncorrelated errors. Then a simple averaging method can effectively reduce the average error of a base model. However, with a sufficiently large pool of base GNN models in the ensemble, it is unrealistic to assume that the errors from these individual base learners are completely uncorrelated. The worst scenario represents another end of the spectrum: the errors are duplicated for all base GNN models, which we aim to avoid. While our focus is on overfitting reduction with substructure and subfeature sampling, 
model diversity by exploring different model architectures and post-processing diversity improvement~\cite{wu2021boosting} for GNNs can be considered in future work. 

\begin{figure*}[t]
  \centerline{\includegraphics[scale=.45]{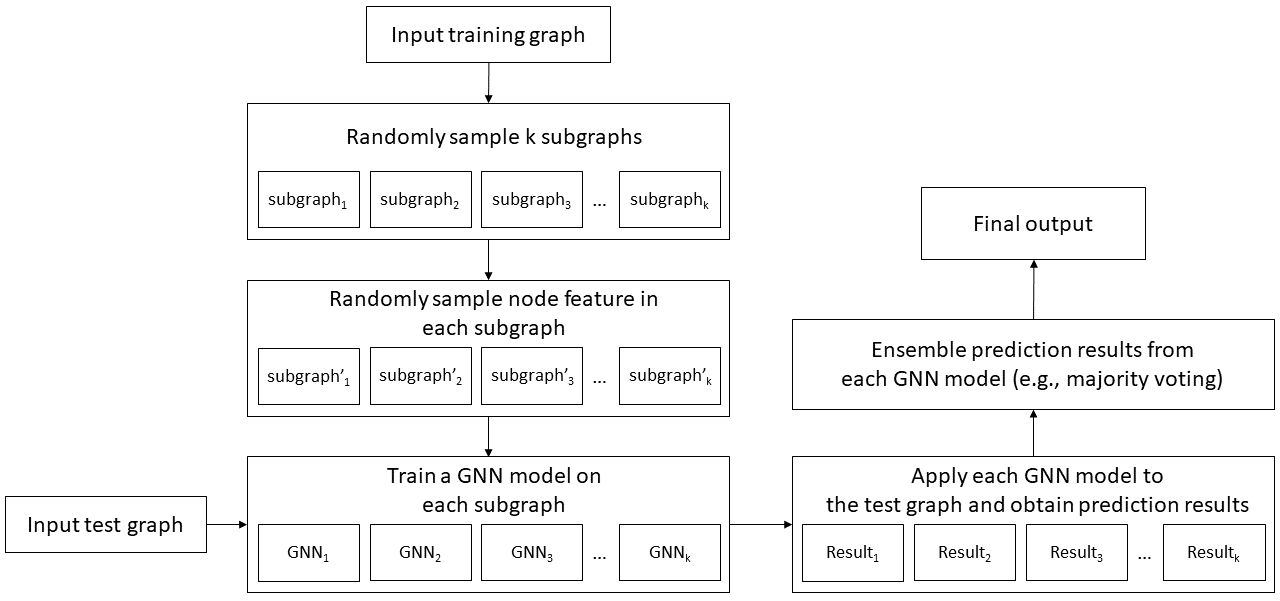}}
\caption{An overview of GNN-Ensemble for node classification. For the transductive setting, the input training graph and input test graph form the full graph.}
\label{fig:GNN-Ensemble} 
\end{figure*}

Our approach is fundamentally different from the existing GNN ensemble literature~\cite{kosasih2021graph,chakravarty2020learning} and implementations\footnote{"Ensemble models for node classification", \url{https://stellargraph.readthedocs.io/}}. These methods train multiple GNN models on the full set of the graph and feature space and perform ensemble learning based on model stacking. In contrast, our method trains the base GNN model on the subgraphs which are sampled from the power set of the graph structure and node feature space. \textbf{Figure~\ref{fig:GNN-Ensemble}} illustrates an overview of GNN-Ensemble during training and at inference for the node classification task. During the training pipeline, we first perform substructure sampling and subfeature sampling. Then we train multiple GNN models in each sampled substructure and subfeature space. During the test phase, a test node is submitted to GNN-Ensemble and is evaluated with each base model. Finally, we aggregate these independent predictions and make the final decision on the nodclass label of the test node.
Our method is especially suitable for parallel implementation of both the generation and evaluation of base models in the ensemble.

\subsection{The Discriminant Function}

Similar to the relation of random forests and decision tree~\cite{ho1995random}, the theoretical foundation of the discriminant power for GNN-Ensemble is built on the theory of stochastic modeling. 

For a given node $x$, let $g_j(x)$ be the GNN trained on the $j$th sampled subgraph $G_j$ $(j=1,2,...,k)$. Let the posterior probability that $x$ is predicted as class $c$  $(c=1,2,...,s)$ be denoted as $P(c|g_j(x))$.

$$P(c|g_j(x)) =\frac{P(c, g_j(x))}{\sum \nolimits_{c=1}^s P(c,g_j(x))}$$
can be estimated by the fraction of class $c$ nodes over all nodes that are assigned to the decision region of $g_j(x)$. 

For an unseen node for prediction, our method averages over the posterior probabilities $P(c|g_j(x))$ that are conditioned on each of the independently trained GNN models. Accordingly, the discriminant function is defined as: $$g_j(x)=\frac{1}{k}\sum\nolimits_{j=1}^{k}P(c|g_j(x)).$$
The decision rule is to assign $x$ to class $c$ for which
$g_j(x)$ is the maximum.
Geometrically, each GNN model defines a neighborhood around the decision region of that node in the chosen subfeature space and substructure space. 

By averaging over the posterior probabilities in these neighborhoods (decision regions), the discriminant function approximates the posterior probability for a given input in the original structure and feature space. The averaging of GNNs lowers the overall variance and prediction error. Similar to random forests~\cite{breiman2001random}, according to the Law of Large Numbers~\cite{kleinberg1996overtraining}, the variance of generalization error is decreasing to zero in GNN-Ensemble when more GNNs are added to the ensemble. Therefore, GNN-Ensemble is an effective method to overcome the overfitting problem. 

\begin{table*}[t]
\caption{Time and Space complexity analysis. $\alpha^*$ denotes the percentage of edges preserved due to the subgraph sampling.}
\begin{minipage}{0.39\linewidth}
\centering
\subcaption{Mathematical notations and explanations. }
\scalebox{1.0}{
\begin{tabular}{|c|c|}
\hline
$l$    & \# layers in GNN      \\ \hline
$k$ & \# base models in ensemble learning  \\ \hline
$n$ & \# nodes in the graph \\ \hline
$m$ & \# edges in the graph \\ \hline
$\alpha$ & \% subgraph     \\ \hline
$d$    & \# features   \\ \hline
$\beta$ & \% subfeature        \\ \hline
$b$    & batch size    \\ \hline
$r$ & \# sampled neighbors \\ \hline
\end{tabular}
}
\label{table:complexity_notation}
\end{minipage}
\begin{minipage}{0.59\linewidth}
\centering
 \hspace{0.2cm}
\subcaption{Complexity.}
\scalebox{0.99}{
{
\begin{tabular}{|cc|c|c|}
\hline
\multicolumn{2}{|c|}{}                                               & baseline    & GNN-Ensemble           \\ \hline
\multicolumn{1}{|c|}{\multirow{2}{*}{GraphSage}}  & time   & $O(r^lnd^2)$    & $O(kr^l \alpha n (\beta d)^2)$      \\ \cline{2-4} 
\multicolumn{1}{|c|}{}                            & space  & $O(br^ld+ld^2)$ &   $O(k(br^l\beta d+l(\beta d)^2)$   \\ \hline
\multicolumn{1}{|c|}{\multirow{2}{*}{FastGCN}}    & time   & $O(rlnd^2)$    & $O(krl \alpha n (\beta d)^2)$     \\ \cline{2-4} 
\multicolumn{1}{|c|}{}                            & space & $O(brld+ld^2)$ & $O(k(brl\beta d+l(\beta d)^2))$      \\ \hline
\multicolumn{1}{|c|}{\multirow{2}{*}{ClusterGCN}} & time   & $O(lmd+lnd^2)$ & $O(k(l \alpha^* m \beta d+l \alpha n (\beta d)^2))$ \\ \cline{2-4} 
\multicolumn{1}{|c|}{}                            & space  & $O(bld+ld^2)$  & $O(k(bl\beta d+l(\beta d)^2))$     \\ \hline
\end{tabular}
}}
\label{table:complexity_complexity}
\end{minipage}
\label{table:complexity}
\end{table*}

\subsection{Decision Aggregation}

GNN-Ensemble makes the final decision by combining individual predictions from all the base GNN models via a committee consensus method.
With the decisions made by $K$ GNN models separately trained on different substructures of the graph and subfeatures of the node features, these decisions are aggregated using the voting method. The setting of three common voting methods is listed below.

\begin{itemize}
    \item Hard voting: the final decision is determined by the majority vote of the prediction results from all decision-makers. If two classes have the same number of votes, we will compare their average prediction confidence and output the higher one.  
    \item Soft voting: the final decision is determined by the highest prediction confidence in all classes by averaging the precision vector of all decision-makers. Unlike hard voting which is based on the decision made by the decision maker, soft voting averages the confidence (prediction probability vector) of the decision makers for decision making.
    \item Weighted voting: the final decision is determined by the weighted combination of decisions. For hard voting, this can be done by assigning weight to the single model based on its past accuracy while equal weights are assigned to all single base GNN models at the beginning. For soft voting, the weight can be given in proportion to the prediction confidence.
\end{itemize}
Given different decision aggregation approaches, 
we observe that no single voting method can consistently outperform the others. Specifically, we find that hard voting and soft voting only have a slight difference: $\pm 0.2\%$. Therefore, we only report the results based on soft voting for brevity.

\subsection{Complexity Analysis}

We analyze and compare the time and space complexity of GNN-Ensemble with different GNN models as the backbone, based on the complexity analysis from ~\cite{chiang2019cluster}. For better readability, we provide \textbf{Table~\ref{table:complexity_notation}} to list all the notations in the complexity analysis. 
With GraphSage~\cite{hamilton2017inductive}, fastGCN~\cite{chen2018fastgcn}, and ClusterGCN~\cite{chiang2019cluster} as examples, \textbf{Table~\ref{table:complexity_complexity}} shows the complexity analysis. We make the following two observations: 
\begin{enumerate}
    \item While we have $k$ base GNN models in the ensemble and the time complexity is multiplied by a $k$ factor, GNN-Ensemble can run training and evaluation in parallel.
    \item By simply stacking multiple GNN models, GNN-Ensemble will introduce additional memory overhead compared to the single model setting, just like all other ensemble models. However,  with a smaller $\alpha$ subgraph percentage and $\beta$ subfeature percentage, the overall time and space complexity can be significantly reduced.
\end{enumerate}

\section{Experiment}

In this section, we will discuss and analyze the performance of the proposed GNN-Ensemble method. We conduct thorough experiments to validate the superiority of our method from three aspects: (i) improvement of the prediction accuracy; (ii) alleviation of overfitting and (iii) robustness against adversarial attacks. Before diving into experimental results and analysis, we start with the experimental setup.

\begin{table}[t]
\centering
\caption{Benchmark datasets and settings.}
\scalebox{0.99}{
{
\begin{tabular}{|c|ccc|c|}
\hline
            & \multicolumn{1}{c|}{Cora} & \multicolumn{1}{c|}{PubMed} & PPI       & Reddit    \\ \hline
\# nodes    & \multicolumn{1}{c|}{2708} & \multicolumn{1}{c|}{19717}  & 56928     & 232965    \\ \hline
\# edges    & \multicolumn{1}{c|}{5429} & \multicolumn{1}{c|}{44338}  & 1587264   & 114615892 \\ \hline
\# features & \multicolumn{1}{c|}{1433} & \multicolumn{1}{c|}{500}    & 50        & 602       \\ \hline
\# classes  & \multicolumn{1}{c|}{7}    & \multicolumn{1}{c|}{3}      & 121, multiple & 41        \\ \hline
\# training & \multicolumn{1}{c|}{140}  & \multicolumn{1}{c|}{60}     & 44809     & 153431    \\ \hline
\# testing  & \multicolumn{1}{c|}{1000} & \multicolumn{1}{c|}{1000}   & 5525      & 55703     \\ \hline
type        & \multicolumn{2}{c|}{transductive}                                   & \multicolumn{2}{c|}{inductive} \\ \hline
\end{tabular}
}}
\label{table:datasets}
\end{table} 

\subsection{Experiment Setup}
In this section, we briefly describe the experimental setup, including datasets, evaluation metrics, and baseline methods. 

\noindent\textbf{Dataset.}
We evaluate the proposed GNN-Ensemble method on four benchmark graph datasets: two of a small scale, i.e., Cora and PubMed, and two of a larger scale, i.e., PPI and Reddit. Cora and PubMed are citation datasets. Protein-protein interaction (PPI) dataset is a multi-class classification dataset and Reddit is a large social network dataset. Detailed information of these datasets are shown in \textbf{Table~\ref{table:datasets}}.
For the split of training and test data, we follow the original setting  in~\cite{kipf2016semi}. 
Cora and PubMed datasets are used for the transductive learning while PPI and Reddit datasets are used for the inductive learning. For all four datasets, we perform the node classification task.

\noindent\textbf{Evaluation metrics. }
Due to the imbalance of classes in the datasets, we use micro-f1 instead of accuracy for measurement. F1 score is the harmonic mean of precision and recall: 
\[
F1=\frac{2*\text{precision}*\text{recall}}{\text{precision}+\text{recall}}.
\]
Precision is the ratio $\frac{tp}{(tp+fp)}$ of correctly identified positive samples in all predicted positive samples, where $tp$ is the number of true positives and $fp$ is the number of false positives.
Recall is the ratio $\frac{tp}{(tp+fn)}$ of correctly identified positive samples in all observed positive instances, where $fn$ is the number of false negatives.

\noindent\textbf{Baseline methods. }
We use GraphSage~\cite{hamilton2017inductive} as the primary backbone in the evaluation and plug into our method FastGCN~\cite{chen2018fastgcn}, ClusterGCN~\cite{chiang2019cluster}, GraphSAINT~\cite{zeng2019graphsaint,fu2021mimosa}, and GAT~\cite{velivckovic2017graph} as a demonstration of universal add-on. We take these models from the PyTorch Geometric package\footnote{\url{https://github.com/pyg-team/pytorch_geometric}}. and use their default hyperparameter settings in the package. In particular, for GraphSage, the model has three layers with a hidden dimension of 256.

\subsection{Accuracy Improvement}

In this set of experiments, we first show the accuracy improvement of GNN-Ensemble with subfeature sampling on the full graph structure (i.e., no substructure sampling). GraphSage is used as the base GNN model. We then extend the GNN-Ensemble method with the best parameter setting to FastGCN, ClusterGCN, GraphSAINT, and GAT. We consider a single GraphSage model trained with the full graph and all node features as the baseline. We randomly sample 10\%, 30\%, 50\%, and 70\% of the original node features, as well as keep the full set of original features (i.e., 100\%). Meanwhile, we build a massive amount of GNN base models for the ensemble. Empirically, we select the number of base models as 100. \textbf{Table~\ref{table:graphsage_subfeature}} shows the result with node feature sampling on the full graph structure. We make four observations below:
\begin{itemize}
    \item When the sampling percentage of node features is 100\% (i.e., using the full set of original features and full graph structure), GNN-Ensemble stacks 100 GraphSage models. Its classification accuracy on the Cora dataset is 0.797, while the accuracy of a single GraphSage model is 0.789. Simply stacking 100 GNN models can already improve the performance and the randomness introduced by the node subfeature sampling can further boost the improvement. 
    \item GNN-Ensemble improves the performance when there is rich node feature information in the dataset. Specifically, GNN-Ensemble achieves about more than 1\% improvement on Cora, which has 1433 node features, and 0.4\% for PubMed with 500 node features. 
    \item Although GNNs do not directly act on the hyperplane of each node feature like the decision trees in random forests, GNN-Ensemble can benefit from sampling node features. Empirically, the boost of accuracy is maximized when the subfeature sampling rate falls into a certain range, i.e., 30\% $\sim$ 70\% of total features in the measurement.
    \item Even with as small as 10\% of the total features, the resulting GNN-Ensemble model can be as nearly accurate as the model with full features. 
\end{itemize}

\begin{table}[t]
\centering
\caption{GNN-Ensemble with GraphSage under the full graph and different percentages of subfeature.}
\scalebox{0.99}{
{
\begin{tabular}{|c|c|ccccc|}
\hline
GraphSage     & baseline & \multicolumn{5}{c|}{GNN-Ensemble: 100 model}                                                                                \\ \hline
subfeature \% & 100                & \multicolumn{1}{c|}{10}     & \multicolumn{1}{c|}{30}    & \multicolumn{1}{c|}{50}   & \multicolumn{1}{c|}{70}   & 100     \\ \hline
Cora          & 0.789            & \multicolumn{1}{c|}{0.792} & \multicolumn{1}{c|}{0.801} & \multicolumn{1}{c|}{\textbf{0.804}} & \multicolumn{1}{c|}{0.803} & 0.797 \\ \hline
PubMed        & 0.781            & \multicolumn{1}{c|}{0.782} & \multicolumn{1}{c|}{\textbf{0.785}} & \multicolumn{1}{c|}{0.784} & \multicolumn{1}{c|}{0.784} & 0.781 \\ \hline
\end{tabular}
}}
\label{table:graphsage_subfeature}
\end{table} 

Then we look at the combined impact of the substructure sampling and subfeature sampling. \textbf{Table~\ref{table:graphsage_subgraph}} shows the results of different percentages of subfeatures with 30\% and 70\% sampled graph structures when training a base GNN model in GNN-Ensemble. That is, we keep only 30\% and 70\% of the nodes in the full graph and their corresponding edges, and vary the sampling percentage of node features at the same time. We make three observations:
\begin{itemize}
\item Compared to Table~\ref{table:graphsage_subfeature}, when 70\% of the graph nodes are randomly sampled, for the Cora dataset, GNN-Ensemble consistently outperforms GNN-Ensemble with only subfeature sampling. GNN-Ensemble with both substructure and subfeature sampling is able to improve the performance over the single model baseline when we maintain a certain percentage of the graph structure for each model. 
\item The size of the subgraph will affect the overall performance even with full node features. Empirically, we find that the subgraph size of 70\% can be a reasonable setting for preserving the needed structural information. Only keeping 30\% of the nodes would remove too much graph information and therefore lower the f1 score.
\item  Another merit of GNN-Ensemble is privacy protection.
Suppose each party holds a subgraph (with subfeatures), without access to the full graph due to privacy restrictions, we can apply GNN-Ensemble to train a GNN at each party and then ensemble them for final predictions. This is similar to the vertical federated learning setup~\cite{yang2019federated}.
\end{itemize}

\begin{table}[t]
\caption{GNN-Ensemble with GraphSage under different percentages of subgraph and subfeature.}
\begin{minipage}{0.99\linewidth}
\centering
\subcaption{70\% subgraph}
\scalebox{0.95}{
\small{
\begin{tabular}{|c|c|ccccc|}
\hline
GraphSage     & baseline & \multicolumn{5}{c|}{GNN-Ensemble: 100 model}                                                                                \\ \hline
subfeature \% & 100                & \multicolumn{1}{c|}{10}     & \multicolumn{1}{c|}{30}    & \multicolumn{1}{c|}{50}    & \multicolumn{1}{c|}{70}    & 100   \\ \hline
Cora          & 0.789            & \multicolumn{1}{c|}{0.792} & \multicolumn{1}{c|}{0.801}  & \multicolumn{1}{c|}{\textbf{0.805}} & \multicolumn{1}{c|}{\textbf{0.805}}  & 0.802 \\ \hline
PubMed        & 0.781            & \multicolumn{1}{c|}{0.782} & \multicolumn{1}{c|}{0.784} & \multicolumn{1}{c|}{\textbf{0.785}} & \multicolumn{1}{c|}{0.782} & 0.781 \\ \hline
\end{tabular}
}}
\label{table:graphsage_subgraph_70}
\end{minipage}
\begin{minipage}{0.99\linewidth}
\hspace{0.2cm}
\centering
\subcaption{30\% subgraph}
\scalebox{0.95}{
\small{
\begin{tabular}{|c|c|ccccc|}
\hline
GraphSage     & baseline & \multicolumn{5}{c|}{GNN-Ensemble: 100 model}                                                                                \\ \hline
subfeature \% & 100                & \multicolumn{1}{c|}{10}     & \multicolumn{1}{c|}{30}    & \multicolumn{1}{c|}{50}    & \multicolumn{1}{c|}{70}    & 100   \\ \hline
Cora          & 0.789            & \multicolumn{1}{c|}{0.768} & \multicolumn{1}{c|}{0.780} & \multicolumn{1}{c|}{0.781} & \multicolumn{1}{c|}{0.784} & \textbf{0.788} \\ \hline
PubMed        & 0.781            & \multicolumn{1}{c|}{0.760} & \multicolumn{1}{c|}{\textbf{0.779}} & \multicolumn{1}{c|}{0.776} & \multicolumn{1}{c|}{0.768} & 0.772 \\ \hline
\end{tabular}
}}
\label{table:graphsage_subgraph_30}
\end{minipage}
\label{table:graphsage_subgraph}
\end{table}

\begin{table*}[t]
\centering
\caption{GNN-Ensemble with different types of GNN models as the base model: aggregate 100 base models, each of which is trained with 70\% subgraph and 50\% subfeatures.}
\scalebox{0.99}{
\small{
\begin{tabular}{|c|cc|cc|cc|cc|}
\hline
\multirow{2}{*}{} & \multicolumn{2}{c|}{Cora}                  & \multicolumn{2}{c|}{PubMed}                & \multicolumn{2}{c|}{PPI}                   & \multicolumn{2}{c|}{Reddit}                \\ \cline{2-9} 
                  & \multicolumn{1}{c|}{baseline} & GNN-Ensemble & \multicolumn{1}{c|}{baseline} & GNN-Ensemble & \multicolumn{1}{c|}{baseline} & GNN-Ensemble & \multicolumn{1}{c|}{baseline} & GNN-Ensemble \\ \hline
Fast-GCN          & \multicolumn{1}{c|}{0.674}    & \textbf{0.692}      & \multicolumn{1}{c|}{0.718}    & \textbf{0.735}      & \multicolumn{2}{c|}{cannot converge}       & \multicolumn{1}{c|}{0.937}    & \textbf{0.951}      \\ \hline
Cluster-GCN       & \multicolumn{1}{c|}{0.812}    & \textbf{0.831}      & \multicolumn{1}{c|}{0.768}    & \textbf{0.774}      & \multicolumn{1}{c|}{0.982}    & \textbf{0.989}      & \multicolumn{1}{c|}{0.954}    & \textbf{0.967}      \\ \hline
GraphSage         & \multicolumn{1}{c|}{0.789}    & \textbf{0.805}      & \multicolumn{1}{c|}{0.781}    & \textbf{0.785}      & \multicolumn{1}{c|}{0.598}    & \textbf{0.614}      & \multicolumn{1}{c|}{0.953}    & \textbf{0.965}      \\ \hline
GraphSaint        & \multicolumn{1}{c|}{0.787}    & \textbf{0.806}      & \multicolumn{1}{c|}{0.751}    & \textbf{0.764}      & \multicolumn{1}{c|}{0.98}     & \textbf{0.989}      & \multicolumn{1}{c|}{0.964}    & \textbf{0.968}     \\ \hline
GAT               & \multicolumn{1}{c|}{0.832}    & \textbf{0.849}      & \multicolumn{1}{c|}{0.779}    & \textbf{0.786}      & \multicolumn{1}{c|}{0.988}    & \textbf{0.993}      & \multicolumn{1}{c|}{0.967}    & \textbf{0.979}      \\ \hline
\end{tabular}
}}
\label{table:generalization}
\end{table*} 
Next, we investigate the generalization of GNN-Ensemble using other types of GNN models as backbones. Based on the aforementioned results, we use the best setting of 70\% of sampled
graph structures and 50\% of sampled node features to train the GNN base model in the ensemble. In practice, the sampling percentages of structure and feature are hyper-parameters and can be optimized through cross-validation. \textbf{Table~\ref{table:generalization}} shows the result. We make three observations.
\begin{itemize}
\item  GNN-Ensemble is not limited to GraphSage and can be applied to many other types of GNN models, such as FastGCN, ClusterGCN, GraphSaint, and GAT. It can be further extended beyond those GNN models provided in the PyTorch Geometric package and can serve as a universal add-on.
\item GNN-Ensemble is not limited to the transductive setting, such as classifications on Cora and PubMed. It can be applied to the inductive setting as well, such as classifications on PPI and Reddit. In our experiment, GNN-Ensemble is able to improve the f1 score 1\% $\sim$ 2\% under the transductive setting and around 1\% under the inductive setting. 
\item While the baseline model may converge to different local optimums, all types of sub-optimal results can be improved with GNN-Ensemble. For example, although the f1 score of GraphSage on PPI is 0.598, GNN-Ensemble improves its performance to 0.614. The GAT on PPI is well converged to 0.988, but GNN-Ensemble can still increase its f1 score to 0.993.
\end{itemize}

\subsection{Overfitting Reduction}

\begin{table}[t]
\centering
\caption{Overfitting reduction measurement on original and reversed Reddit dataset: GraphSage is used with two settings of hidden parameters: 256 and 2048. GNN-Ensemble aggregates 100 base models, each of which is trained with 70\% subgraph and 50\% subfeature.}
\scalebox{0.99}{
\small{
\begin{tabular}{|c|c|cc|cc|}
\hline
\multirow{2}{*}{}        & \multirow{2}{*}{\# embedding} & \multicolumn{2}{c|}{baseline}           & \multicolumn{2}{c|}{GNN-Ensemble}          \\ \cline{3-6} 
                         &                               & \multicolumn{1}{c|}{training} & testing & \multicolumn{1}{c|}{training} & testing \\ \hline
\multirow{2}{*}{vanilla} & 256                           & \multicolumn{1}{c|}{0.9766}   & 0.953   & \multicolumn{1}{c|}{0.9648}   & 0.965   \\ \cline{2-6} 
                         & 2048                          & \multicolumn{1}{c|}{0.9789}   & 0.954   & \multicolumn{1}{c|}{0.9781}   & 0.970   \\ \hline
\multirow{2}{*}{reverse} & 256                           & \multicolumn{1}{c|}{0.9993}   & 0.95    & \multicolumn{1}{c|}{0.9987}   & 0.966   \\ \cline{2-6} 
                         & 2048                          & \multicolumn{1}{c|}{0.9999}   & 0.951   & \multicolumn{1}{c|}{\textbf{0.9999}}   & \textbf{0.973}   \\ \hline
\end{tabular}
}}
\label{table:overfitting}
\end{table} 

Overfitting is more often to occur when training on smaller datasets with more complex models. We demonstrate that GNN-Ensemble can effectively reduce the  overfitting issue. We conduct experiments on the Reddit data with vanilla and reversed training and test split setup. Specifically, in the vanilla setup, we use the original training and test data (153,431 training nodes and 55,703 test nodes, as shown in Table~\ref{table:datasets}), where the size of training data is significantly larger than that of test data. While in the reversed setup, we reverse the training and test data, that is, 55,703 training nodes and 153,431 test nodes, so that the training data becomes much smaller. In addition, we purposely increase the number of embeddings in the hidden layer of GraphSage from 256 to 2048, in order to make a heavily parameterized GNN model. As a result, in the reversed setup, overfitting becomes very evident since there are many more training parameters in the GNN model with much less training data. \textbf{Table~\ref{table:overfitting}} shows the result and we make two observations:
\begin{itemize}
    \item GNN-Ensemble is able to reduce overfitting. Specifically, the difference between training and test performance of GNN-Ensemble is smaller than that of the single model baseline. Such observation holds for both vanilla and reversed setups as well as for both 256 and 2048 dimensional embedding. The performance of GNN-Emsemble on test data is consistently better than that of the baseline. 
   \item   GNN-Ensemble can reduce overfitting more effectively when  the complexity of the model is higher. Under the reversed setting, when the dimension of embedding is 2048, GNN-Ensemble increases the f1 score by 2\% in the test dataset.
\end{itemize}

\subsection{Adversarial Robustness}

At last, we measure the adversarial robustness of GNN-Ensemble. We conduct the experiment using DeepRobust\footnote{https://github.com/DSE-MSU/DeepRobust} and consider four graph evasion attacks~\cite{jin2003adversarial}: RL-S2V~\cite{dai2018adversarial}, PGD~\cite{xu2019topology}, GF-Attack~\cite{chang2020restricted}, and IG-FGSM~\cite{wu2019adversarial}. All four attacks perturb the input graph by adding and deleting edges. IG-FGSM, in addition, modifies the features as well. We set the maximum perturbation threshold to 10\%, that is, allow the attacker to add or delete at most 10\% of the edges. To satisfy the imperceptibility requirement of adversarial attacks, the attacker can only modify such a limited percentage of nodes or edges. We compare GNN-Ensemble with three representative adversarial defense approaches: GCN-SVD~\cite{entezari2020all}, GCN-Jaccard~\cite{wu2019adversarial} and ProGNN~\cite{jin2020graph}. These defense approaches are applied in their default settings in DeepRobust.
This set of experiments are conducted on Cora and PubMed under the transductive setting. 
\textbf{Table~\ref{table:adversarial}} shows the adversarial robustness results. We make three observations: 

\begin{table}[t]
\centering
\caption{Adversarial robustness measurement: GNN-Ensemble aggregates 100 base models, each of which is trained with 70\% subgraph and 50\% subfeature.}
\scalebox{0.90}{
{
\begin{tabular}{|cc|c|c|c|c|c|}
\hline
\multicolumn{2}{|c|}{}                                      & benign & RL-S2V & PGD   & GF-attack & IG-FGSM \\ \hline
\multicolumn{1}{|c|}{\multirow{5}{*}{\rotatebox{90}{Cora}}}   & baseline    & 0.789  & 0.712  & 0.72  & 0.726     & 0.025   \\ \cline{2-7} 
\multicolumn{1}{|c|}{}                        & GCN-SVD     & 0.756  & 0.683  & 0.691 & 0.678     & 0.566   \\ \cline{2-7} 
\multicolumn{1}{|c|}{}                        & GCN-Jaccard & 0.773  & 0.741  & 0.745 & 0.737     & 0.736   \\ \cline{2-7} 
\multicolumn{1}{|c|}{}                        & ProGNN      & 0.782  & 0.745  & 0.738 & 0.741     & 0.739   \\ \cline{2-7} 
\multicolumn{1}{|c|}{}                        & GNN-Ensemble  & \textbf{0.805}  & \textbf{0.769}  & \textbf{0.773} & \textbf{0.761}     & \textbf{0.775}   \\ \hline
\multicolumn{1}{|c|}{\multirow{5}{*}{\rotatebox{90}{PubMed}}} & baseline    & 0.781  & 0.709  & 0.715 & 0.714     & 0.011   \\ \cline{2-7} 
\multicolumn{1}{|c|}{}                        & GCN-SVD     & 0.738  & 0.688  & 0.684 & 0.689     & 0.557   \\ \cline{2-7} 
\multicolumn{1}{|c|}{}                        & GCN-Jaccard & 0.774  & 0.745  & \textbf{0.746} & 0.743     & 0.725   \\ \cline{2-7} 
\multicolumn{1}{|c|}{}                        & ProGNN      & 0.776  & 0.723  & 0.730  & 0.731     & 0.734   \\ \cline{2-7} 
\multicolumn{1}{|c|}{}                        & GNN-Ensemble  & \textbf{0.785}  & \textbf{0.751}  & 0.744 & \textbf{0.748}     & \textbf{0.759}   \\ \hline
\end{tabular}
}}
\label{table:adversarial}
\end{table} 

\begin{itemize}
\item For all four adversarial attacks on the graph, GNN-Ensemble consistently and significantly outperforms the four defense methods on both datasets. Since the attacker is limited to modifying at most 10\% of the edges, the majority of the nodes and edges remain unchanged during the adversarial attack. Since GNN-Ensemble explores the structural information of the graph and uses multiple subgraphs of the original graph for learning and prediction, subgraphs containing the victim node will be less likely to be impacted by the maliciously modified edges. Therefore, GNN-Ensemble can achieve better classification performance on the victim nodes in an adversarially sabotaged graph.
\item GNN-Ensemble is very effective in mitigating the adversarial effect of IG-FGSM, where attacks occur by
adding and deleting edges as well as modifying features. GNN-Ensemble trains the base GNN models on sampled substructures and subfeatures of the original graph. The adversarial effect is much less eminent in the sampled subspaces. Therefore, GNN-Ensemble is very robust to adversarial attacks.
\item Due to the different optimization goals under every possible combination of substructure and subfeatures, it now becomes very difficult to make adversarial examples via optimizing the graph in favor of the attackers. Therefore, different base models in GNN-Ensemble can hardly be jointly optimized to perform the cross-model attack~\cite{athalye2018obfuscated}.
\end{itemize}

\section{Conclusion}
We propose a new graph learning method called GNN-Ensemble to construct an ensemble of random decision graph neural networks. GNN-Ensemble consists of multiple base models that are trained with randomly selected substructures in the topological space and subfeatures in the node feature space. The discriminant function 
of GNN-Ensemble approximates the posterior probability for a given input in the original topological and feature space. GNN-Ensemble is easily parallelized, where each base model can be trained and perform inference independently. In addition, different types of GNN models can be used as the base model in the ensemble. Extensive experimental results on four real-world benchmark graph datasets show that GNN-Ensemble consistently outperforms all the baselines on various node classification tasks. Similar to random forests, GNN-Ensemble is also very effective in overcoming the overfitting problem. Last but not the least,  GNN-Ensemble significantly improves the adversarial robustness against attacks on single GNN model. 



\bibliographystyle{IEEEtran}
\bibliography{reference}

\end{document}